# Softening Fuzzy Knowledge Representation Tool with the Learning of New Words in Natural Language

Mohamed-Nazih Omri
*ERPAH-FST & DMI-IPEIM,*
*Route de Kairouan, 5019 Monastir.*
*Tel. 216 3 500 273, Fax. 216 3 500 512*
Nazih.Omri@ipeim.rnu.tn

**Abstract**

The approach described here allows using membership function to represent imprecise and uncertain knowledge by learning in Fuzzy Semantic Networks. This representation has a great practical interest due to the possibility to realize on the one hand, the construction of this membership function from a simple value expressing the degree of interpretation of an Object or a Goal as compared to an other and on the other hand, the adjustment of the membership function during the apprenticeship. We show, how to use these membership functions to represent the interpretation of an Object (respectively of a Goal) user as compared to an system Object (respectively to a Goal). We also show the possibility to make decision for each representation of an user Object compared to a system Object. This decision is taken by determining decision coefficient calculates according to the nucleus of the membership function of the user Object.

## 1. Introduction

Users do not learn through written instructions. Learning how to use a new technological system is mainly an exploratory activity. Exploring learning has shown to improve the abilities constructing to successful error handling and discovering and eventually constructing correct knowledge about the query, but exploratory activity frequently leads to experience uninterested states or not reach the interested state Goal. Users need assistance not only to avoid errors, but to understand how the system interprets their commands and How and Why to act in order to reach their System's Goal [Richard & al. 93].

In order to respond to a query, an executive assistant might know very precisely the Goal the user has in mind, which means an Object in a given state (the properties of the Object being transformed). Moreover, even when Goals are fairly well defined, it is often necessary to think about superordinate Goals.

A novice user asks the expert operator about how to execute the action 'how to Gum Word ?'. The problem that is under investigation is how to match the content of this query to their corresponding items. Precisely, how to match the Goal 'Gum' to its corresponding item in the set of standard nominations {EraseWithMenu, EraseWithKey, CutWithMenu, Copy...}. The item 'Word' in the user's query is interpreted easily by the system because it corresponds to the word Object's class in the semantic net work[Omri & al. 99a], but the Goal 'Gum' is not an element of the Goal's standard set and then with an interactive dialogue, the system tries to identify the unknown Goal 'Gum'. It presents the set of Goals that may be applied on the Object 'Word', asks the user to associate a possibility degree taken between 0 and 1 for each element of the standard set. This degree quantify the opinion of the users about the ideal system Goal and allows us to construct this membership function from a simple value expressing the degree of interpretation of an Object or a Goal as compared to an other and on the other hand, the manner to adjust it during the apprenticeship. We show also, how to use these functions of membership to represent the interpretation of an user Object (respectively of a Goal) as compared to a system Object (respectively of a Goal).

In our technological system[Omri & al. 95], learning is defined as the system's capacity to interpret an unknown word using the links created between this new word and known words. The main link is provided by the context of the query. When novice's query is confused with an unknown verb (Goal) applied to a known noun denoting either an Object in the ideal user's Network or an Object in the user's Network, the system infer that this new verb corresponds to one of the known Goal.

## 2. Construction of the membership function

We consider a membership function characterized by a quadrupled $[\gamma, \alpha, \beta, \delta]$, where $\alpha$ and $\beta$ are respectively the inferior and superior stone of the nucleus, $\gamma$ and $\delta$ are respectively the inferior and superior stone of the support of this membership function. The alone value associated by a user to a Goal (respectively an Object) to designate a Goal (respectively an Object) system allows to obtain the departure function membership (Fig. 1) (Fig. 2) of the Goal (respectively of the Object) aimed. The nucleus of this function will be characterized by the value $\alpha$ ($\alpha = \beta$) associated by the user and the support will be delimited according to cases: if $\vartheta < 0.5$ then $\gamma = 0$, $\delta = 2\vartheta$ and if $\vartheta > 0.5$ then $\gamma = 2\vartheta - 1$ and $\delta = 1$.

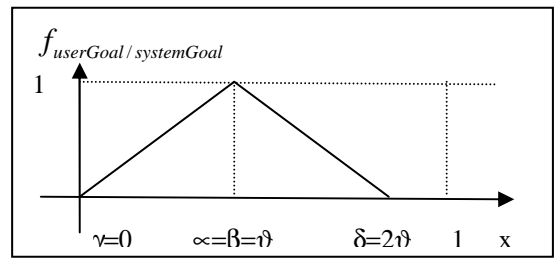

Fig. 1 - Departure membership function: case where $\vartheta \prec 0.5$.

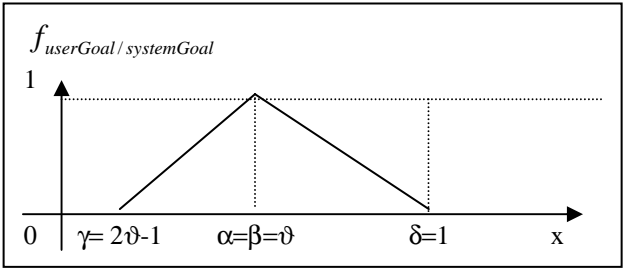

Fig. 2 - Departure membership function: case where $\vartheta \succ 0.5$.

This membership function thus constructed (Fig.1), undergo an adjustment during the manipulation of the Object or the Goal. The different terms that users use to designate an Object or a Goal given system allow to realize this adjustment. This last can be undertaken two different manners: to strengthen the membership function or to weaken it, according to values attributed by the user. The Objective is to represent efficiently the Goal (respectively the Object) user by this membership function according to a Goal (respectively Object) system.

### 2.1. Adjustment of the f's nucleus

The nucleus of the membership function is reduced to the value $\alpha = \beta$. In the course of the first iteration, a user associates the value 0 to the Goal (respectively Object) to represent. The correction that it is necessary to bring to the considered nucleus consists in adjust one stone of the nucleus according to the value $\vartheta$ associated as following.

- If $\vartheta > \alpha$, then $\alpha$ been the inferior stone by the nucleus and the superior stone $\beta$. This last will be adjusted by:

$$\text{Ajus}(\beta) = \frac{\beta + \vartheta}{2}$$

- If $\vartheta < \alpha$, we guards $\beta$ as superior stone and we adjusts the inferior stone $\alpha$ by:

$$\text{Ajus}(\alpha) = \frac{\alpha + \vartheta}{2}$$

The figure (Fig. 3) shows the new value of the nucleus after adjustment.

For two iterations, the coefficient of adjustment is calculates by the next formula:

$$Ajus(\alpha) = \frac{2*\alpha + \vartheta}{3}$$

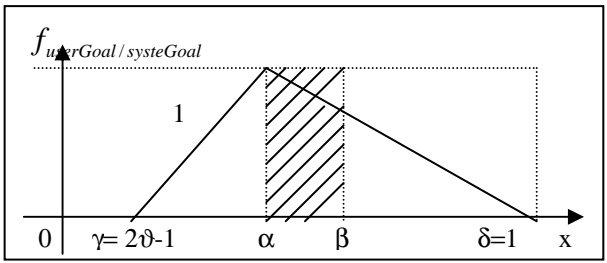

Fig. 3 - Adjustment of the f's nucleus.

This formula holds therefore account the number of adjustment that has undergone the stone has the nucleus. The same principle is applied if the adjustment concerns the stone $\beta$. Consequently, by generalizing the adjustment for a number N of iterations, the formula of the nucleus stone adjustments of f become:

$$Ajus(\alpha_{N+1}) = \frac{N*\alpha_N + \vartheta_{N+1}}{N+1}, \text{ if } \vartheta_{N+1} \leq \frac{\alpha_N + \beta_N}{2}$$

$$Ajus(\beta_{N+1}) = \frac{N*\beta_N + \vartheta_{N+1}}{N+1}, \text{ if } \vartheta_{N+1} \geq \frac{\alpha_N + \beta_N}{2}$$

Where Ajus ($\alpha$) and Ajus ($\beta$) are respectively the new values of $\alpha$ and $\beta$ after adjustment.

We notices that this adjustment undertakes only an alone side both since this depends on the value of has as compared to stones of the nucleus. Two possible cases can happen: the first case produces when $\vartheta$ is inferior to the average of the nucleus ($\vartheta \leq \frac{\alpha + \beta}{2}$) and therefore the adjustment is called to left, the second case produces when $\vartheta$ is superior to this average ($\vartheta \geq \frac{\alpha + \beta}{2}$), in this last case, the adjustment is called to straight.

### 2.2. Adjustment of the f's support

Concerning the adjustment of the f's support, the same reasoning applied on the case of the adjustment of the nucleus, is used in this case. Two possible cases can happen: the case where ($\vartheta \leq \frac{\alpha + \beta}{2}$) and the opposite case, it is had called ($\vartheta \geq \frac{\alpha + \beta}{2}$). In the first case the adjustment concerns the inferior stone by the support and in the opposite case, it concerns the stone superior. For the case of the adjustment of the inferior stone, the expression is given by:

$$Ajus(\gamma_{N+1}) = \frac{N*\gamma_{N+1} + \vartheta_{N+1}}{N+1}, \text{ if } \vartheta_{N+1} \leq \frac{\alpha_N + \beta_N}{2}$$

For the adjustment of the superior stone of the support, the formula above becomes:

Ajus ($\gamma$) and Ajus ($\delta$) are respectively the new values of $\gamma$

$$Ajus(\delta_{N+1}) = \frac{N*\delta_N + \vartheta_{N+1}}{N+1}, \text{ if } \vartheta_{N+1} \geq \frac{\alpha_N + \beta_N}{2}$$

and $\delta$ after adjustment.

The figure 4 presents the membership function after adjustment of its nucleus and its support.

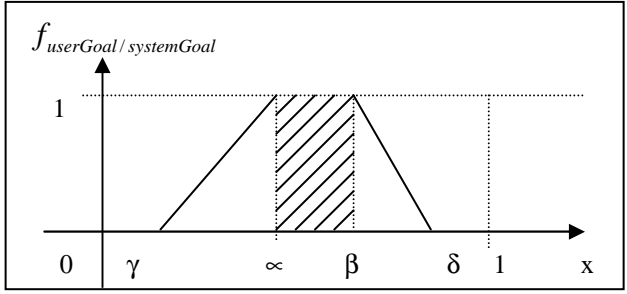

Fig. 4 - Adjustment of the f's support.

**Example 1**

Let 'Substantive' be the liable Object that one seeks to represent and 'Word' the Object ideally considered. Let $\vartheta = 0.7$ the degree of possibility associated by the user to the Object 'Substantive' to designate the Object system 'Word'. A second user associates the value $\vartheta = 0.5$ as the degree of confidence, we have therefore $\vartheta \leq \frac{\alpha+\beta}{2}$ and N = 1 (N been the number of iterations).

Undertaken calculations are detailed in paragraphs a and b following.

a) **adjustment of the nucleus.**

$$Ajus(\alpha_2) = \frac{1*0.7+0.5}{2} = 0.6$$

The new value of α been α = 0.6 and β preserved its value β = 0.7.

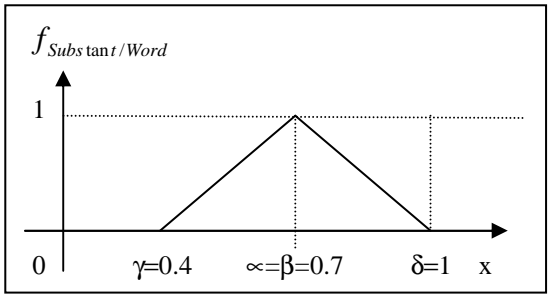

Fig. 5 - associated membership function to 'Substantive'.

b) **adjustment of the support.**

$$Ajus(\gamma_2) = \frac{1*0.4+0.5}{2} = 0.45$$

Knowing that $(\gamma = 0) \leq (\vartheta = 0.5) \leq \frac{\alpha+\beta}{2}$ (old value before adjustment), the undertaken adjustment is always left.

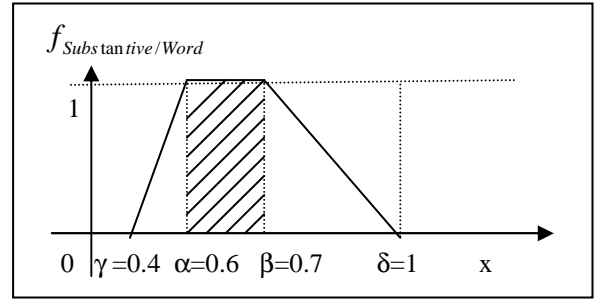

Fig. 6 - Membership function with nucleus adjusted.

The consequent function after adjustment of the nucleus and the support is given by the figure (Fig.7).

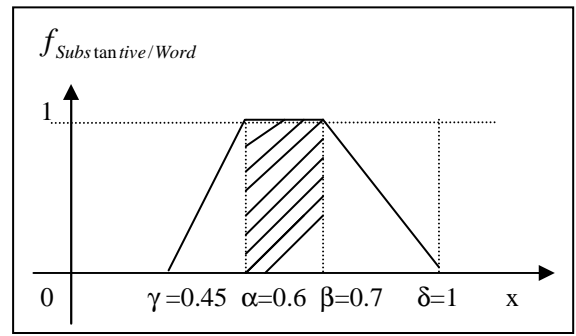

Fig. 7 - Membership function after adjustment.

**3. Coefficient of decision**

In a problem of decision, there exists often a gap between the Object ideally considered and the Object observed. The problem is to measure this gap to tell to what extent an observed Object O satisfies an Object O' theoretical. In our case, an observed Object is represented by a fuzzy subset according to an Object theoretical.

The solution that we have retained consists in determine a coefficient of decision Dc (O/$O_i$) for each representation of an User Object as compared to an Object relative given system to an Object given. This coefficient of decision Dc (O/$O_i$) calculates in function of the nucleus of the membership function of the User Object.

**3.1 Definitions**

**Definition 1.** Let $\alpha_i$ be the inferior stone by the nucleus of the membership function f associated to an User Object $O_i$ as compared to an Object system O and $\beta_i$ the stone superior of the f's nucleus. One defines the coefficient of decision Dc (O/O$_i$) of an User Object O as compared to an Object system O$_i$ by:

$$Dc(O/O_i) = \frac{\alpha_i + 3\beta_i}{4}$$

The choice of this coefficient is made with the next manner : by leaving, in the beginning, the average of the nucleus that forms the totality of most believable values, several functions of membership can present a same average of the nucleus despite that they are different of each other. It is therefore difficult to choose that that represents the better an Object given. To allow an efficient choice, one has introduced the superior stone b of the nucleus in the calculation of the coefficient of decision Dc. Thus, to decide between several Object systems, one calculates the coefficient of decision of each Object. That whose value is the largest is therefore that that represents the better the User Object, and consequently, it is retained among others.

The coefficient allowing to realize the final choice is called coefficient final decision, noted $D_c^f$, and datum by:

**Definition 2.** Let $O_i$, $i \in [1, n]$, be the totality of Object systems whose one knows for each of them, the decision coefficient Dc as compared to a liable Object O. We define the final decision coefficient, noted $D_c^f$, of O as compared to the set $\{O_1, O_2, ..., O_n\}$ by:

$$D_c^f(O/O_i, i \in [1,n]) = \max_{1 \leq i \leq n} D_c(O/O_i) = \max_{1 \leq i \leq n} \frac{\alpha_i + 3\beta_i}{4}$$

**Example 2.**

Let 'Substantive' be the User Object whose one seeks the meaning as compared to the totality of Object system {Character, Word, ChaineofChar.}, noted by {Cha, Mo, Ca}. The corresponding Functions of membership are given respectively by (fig. 8), (fig. 9) and (fig. 10) relative to the User Object 'Substantive'.

One calculates now the different degree $Dc(O/O_i)$, one a:

$$D_c(Subs\tan tif/Ca) = \frac{0.3 + 3*0.6}{4} = 0.525$$

$$D_c(Subs\tan tif/Mo) = \frac{0.2 + 3*0.7}{4} = 0.575$$

$$D_c(Subs\tan tif/Ch) = \frac{0.4 + 3*0.5}{4} = 0.475$$

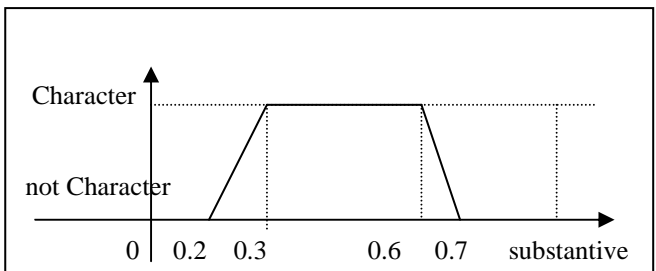

Fig. 8 - Membership function as compared to 'Character'.

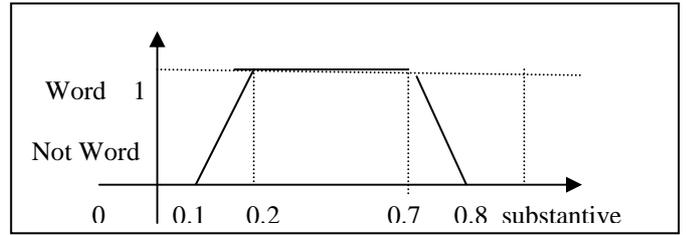

Fig. 9 - Membership function as compared to 'Word'.

Where the final decision coefficient:

$$D_c^f(Subs\tan tive/'Ca, Mo, Ch\}) =$$
$$\max(0.525, 0.575, 0.475) = 0.575$$

We will tell in conclusion that the Object system 'Character' has a closer senses to that the User Object 'Substantive' because it presents the highest coefficient of decision.

Consequently, a query of the type 'How to Select a Substantive?' will have as equivalent request 'How to Select a Character?'. A such request presents no ambiguity, and thus the system can provide a reply to the subject.

**Remark:** The same reasoning is applied for the case of Goals.

The problem that remains to solve is that to use this representation to improve the evolution of the formalism that we have elaborates.

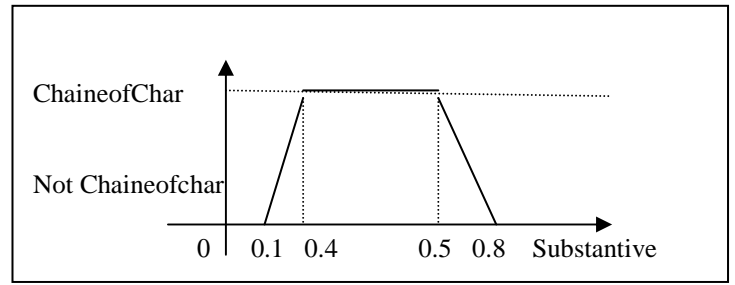

Fig. 10 - Membership function as compared to 'ChaineofChar'.

**4. Conclusion**

With the learning of new words in natural language as the interpretation witch was produced in agreement with the user, the system improves its representation scheme at each experiment with a new user and, in addition, takes advantage of previous discussions with users. Fuzzy semantic Networks of the technological system present three levels or kinds of learning [Omri & al. 95]: we distinguish between states Goals dynamics, states Objects dynamics and states relationship dynamics. In this paper we have described a tool to softening of the fuzzy knowledge representation for the two first kinds of learning, but we think that is possible to represent the third kind of learning witch is the states relations dynamics. Given that each bond of connection is characterized by a couple of value (a degree of possibility and a degree of necessity), the principle will be the even but we have to use two membership functions to represent a bond of connection between two objects or between two purposes. The first function will characterize the degree of possibility and the second the degree of possibility. We think that it would be interesting to strengthen this tool of softening with the notion of similarity between two Objects (respectively two Goals) so as to establish connection between user Object (or Goal) and system Object (or system Goal) in the semantic Net. This makes only increase performances of the system in the course of the identification of user requests.